\RequirePackage[hyphens]{url}
\documentclass[twoside]{article}

\usepackage[accepted]{aistats2023}
\usepackage[round]{natbib}

\usepackage{xcolor}
\usepackage{amsmath}
\usepackage{amssymb}
\usepackage{mathtools}
\usepackage{amsthm}
\usepackage{mathrsfs}
\usepackage[hyphens]{url}
\usepackage{algorithm}
\usepackage{algorithmic}

\begin{document}

%
\newtheorem{theorem}{Theorem}
\newtheorem{corollary}{Corollary}
\newtheorem{lemma}{Lemma}
\newtheorem{definition}{Definition}
\newtheorem{problem}{Problem}
\newtheorem{assumption}{Assumption}
\newtheorem{properties}{Property}
\newtheorem{conjecture}{Conjecture}
\newtheorem{criterion}{Criterion}
\newtheorem{exa}{Example}

\def\nn{\nonumber}
\def\beq{\begin{equation}}
\def\eeq{\end{equation}}
\def\bea{\begin{eqnarray}}
\def\eea{\end{eqnarray}}
\def\ba{\begin{array}}
\def\ea{\end{array}}
\def\defeq{{\stackrel{\Delta}{=}}}

\def\bitem{\begin{itemize}}
\def\eitem{\end{itemize}}
\def\ben{\begin{enumerate}}
\def\een{\end{enumerate}}

\def\eg{{\it e.g., \/}}
\def\etal{{\it et al. \/}}
\def\viz{{\it viz.,\ \/}}
\def\ie{{\it i.e.,\ \/}}
\def\vs{{\it vs. \/}}


\definecolor{bgrd}{rgb}{1,1,1}
\definecolor{gray}{rgb}{0.5,0.5,0.5}
\definecolor{dkr}{rgb}{0.7,0.1,0.2}
\definecolor{dkb}{rgb}{0.1,0.1,0.8}
\def\tcdkr{\textcolor{dkr}}
\def\tcdkb{\textcolor{dkb}}
\def\tcr{\textcolor{red}}
\def\tcb{\textcolor{blue}}
\def\tck{\textcolor{black}}
\def\tcy{\textcolor{yellow}}
\def\tcg{\textcolor{green}}
\def\tcbg{\textcolor{bgrd}}
\def\tcm{\textcolor{magenta}}
\def\tcw{\textcolor{white}}
\def\tcgr{\textcolor{gray}}

\makeatletter
\newdimen{\captionwidth}
\long\def\@makecaption#1#2{%
\captionwidth .9\hsize
\vskip 10pt%
\setbox\@tempboxa\hbox{#1: #2}%
  \ifdim \wd\@tempboxa >\captionwidth%
    \setbox\@tempboxa\hbox{#1:\hspace*{.5em}}%
    \hfil\parbox{\captionwidth}{\raggedright\hangindent \wd\@tempboxa%
    \hangafter=1\unhbox\@tempboxa#2}\hfill%
  \else\centerline{\box\@tempboxa}%
  \fi
}
\makeatother
\def\scalefig#1{\epsfxsize #1\textwidth}

\def\Cov{\mbox{Cov}}
\def\diag{\mbox{diag}}
\def\H{\mbox{\tiny H}}
\def\iid{\stackrel{\mbox{\small i.i.d.}}{\sim}}
\def\LS{\mbox{\tiny LS}}
\def\ML{\mbox{\tiny ML}}
\def\Bayes{\mbox{\tiny Bayes}}
\def\mRe{\mbox{Re}}
\def\mIm{\mbox{Im}}
\def\opt{\mbox{\tiny opt}}
\def\ow{\mbox{otherwise}}
\def\rank{\mbox{rank}}
\def\SNR{\mbox{SNR}}
\def\Tr{\mbox{Tr}}
\def\T{\mbox{\tiny T}}
\def\Var{\mbox{Var}}

\newcommand{\mbbC}{\mathbb{C}}
\newcommand{\mbbE}{\mathbb {E}}
\newcommand{\mbbM}{\mathbb{M}}
\newcommand{\mbbR}{\mathbb{R}}
\newcommand{\mbbV}{\mathbb{V}}

\newcommand{\Amsc}{\mathscr{A}}
\newcommand{\Bmsc}{\mathscr{B}}
\newcommand{\Cmsc}{\mathscr{C}}
\newcommand{\Dmsc}{\mathscr{D}}
\newcommand{\Emsc}{\mathscr{E}}
\newcommand{\Fmsc}{\mathscr{F}}
\newcommand{\Gmsc}{\mathscr{G}}
\newcommand{\Hmsc}{\mathscr{H}}
\newcommand{\Imsc}{\mathscr{I}}
\newcommand{\Jmsc}{\mathscr{J}}
\newcommand{\Kmsc}{\mathscr{K}}
\newcommand{\Lmsc}{\mathscr{L}}
\newcommand{\Mmsc}{\mathscr{M}}
\newcommand{\Nmsc}{\mathscr{N}}
\newcommand{\Omsc}{\mathscr{O}}
\newcommand{\Pmsc}{\mathscr{P}}
\newcommand{\Qmsc}{\mathscr{Q}}
\newcommand{\Rmsc}{\mathscr{R}}
\newcommand{\Smsc}{\mathscr{S}}
\newcommand{\Tmsc}{\mathscr{T}}
\newcommand{\Umsc}{\mathscr{U}}
\newcommand{\Vmsc}{\mathscr{V}}
\newcommand{\Wmsc}{\mathscr{W}}
\newcommand{\Xmsc}{\mathscr{X}}
\newcommand{\Ymsc}{\mathscr{Y}}
\newcommand{\Zmsc}{\mathscr{Z}}

\def\alphabf{\hbox{\boldmath$\alpha$\unboldmath}}
\def\betabf{\hbox{\boldmath$\beta$\unboldmath}}
\def\gammabf{\hbox{\boldmath$\gamma$\unboldmath}}
\def\deltabf{\hbox{\boldmath$\delta$\unboldmath}}
\def\epsilonbf{\hbox{\boldmath$\epsilon$\unboldmath}}
\def\zetabf{\hbox{\boldmath$\zeta$\unboldmath}}
\def\etabf{\hbox{\boldmath$\eta$\unboldmath}}
\def\iotabf{\hbox{\boldmath$\iota$\unboldmath}}
\def\kappabf{\hbox{\boldmath$\kappa$\unboldmath}}
\def\lambdabf{\hbox{\boldmath$\lambda$\unboldmath}}
\def\mubf{\hbox{\boldmath$\mu$\unboldmath}}
\def\nubf{\hbox{\boldmath$\nu$\unboldmath}}
\def\xibf{\hbox{\boldmath$\xi$\unboldmath}}
\def\pibf{\hbox{\boldmath$\pi$\unboldmath}}
\def\rhobf{\hbox{\boldmath$\rho$\unboldmath}}
\def\sigmabf{\hbox{\boldmath$\sigma$\unboldmath}}
\def\taubf{\hbox{\boldmath$\tau$\unboldmath}}
\def\upsilonbf{\hbox{\boldmath$\upsilon$\unboldmath}}
\def\phibf{\hbox{\boldmath$\phi$\unboldmath}}
\def\chibf{\hbox{\boldmath$\chi$\unboldmath}}
\def\psibf{\hbox{\boldmath$\psi$\unboldmath}}
\def\omegabf{\hbox{\boldmath$\omega$\unboldmath}}
\def\Sigmabf{\hbox{$\bf \Sigma$}}
\def\Upsilonbf{\hbox{$\bf \Upsilon$}}
\def\Omegabf{\hbox{$\bf \Omega$}}
\def\Deltabf{\hbox{$\bf \Delta$}}
\def\Gammabf{\hbox{$\bf \Gamma$}}
\def\Thetabf{\hbox{$\bf \Theta$}}
\def\Lambdabf{\mbox{$ \bf \Lambda $}}
\def\Xibf{\hbox{\bf$\Xi$}}
\def\Pibf{{\bf \Pi}}
\def\thetabf{{\mbox{\boldmath$\theta$\unboldmath}}}
\def\Upsilonbf{\hbox{\boldmath$\Upsilon$\unboldmath}}
\newcommand{\Phibf}{\mbox{${\bf \Phi}$}}
\newcommand{\Psibf}{\mbox{${\bf \Psi}$}}

\def\abf{{\bf a}}
\def\bbf{{\bf b}}
\def\cbf{{\bf c}}
\def\dbf{{\bf d}}
\def\ebf{{\bf e}}
\def\fbf{{\bf f}}
\def\gbf{{\bf g}}
\def\hbf{{\bf h}}
\def\ibf{{\bf i}}
\def\jbf{{\bf j}}
\def\kbf{{\bf k}}
\def\lbf{{\bf l}}
\def\mbf{{\bf m}}
\def\nbf{{\bf n}}
\def\obf{{\bf o}}
\def\pbf{{\bf p}}
\def\qbf{{\bf q}}
\def\rbf{{\bf r}}
\def\sbf{{\bf s}}
\def\tbf{{\bf t}}
\def\ubf{{\bf u}}
\def\vbf{{\bf v}}
\def\wbf{{\bf w}}
\def\xbf{{\bf x}}
\def\ybf{{\bf y}}
\def\zbf{{\bf z}}
\def\rbf{{\bf r}}
\def\xbf{{\bf x}}
\def\ybf{{\bf y}}
\def\Abf{{\bf A}}
\def\Bbf{{\bf B}}
\def\Cbf{{\bf C}}
\def\Dbf{{\bf D}}
\def\Ebf{{\bf E}}
\def\Fbf{{\bf F}}
\def\Gbf{{\bf G}}
\def\Hbf{{\bf H}}
\def\Ibf{{\bf I}}
\def\Jbf{{\bf J}}
\def\Kbf{{\bf K}}
\def\Lbf{{\bf L}}
\def\Mbf{{\bf M}}
\def\Nbf{{\bf N}}
\def\Obf{{\bf O}}
\def\Pbf{{\bf P}}
\def\Qbf{{\bf Q}}
\def\Rbf{{\bf R}}
\def\Sbf{{\bf S}}
\def\Tbf{{\bf T}}
\def\Ubf{{\bf U}}
\def\Vbf{{\bf V}}
\def\Wbf{{\bf W}}
\def\Xbf{{\bf X}}
\def\Ybf{{\bf Y}}
\def\Zbf{{\bf Z}}
\def\Ac{{\cal A}}
\def\Bc{{\cal B}}
\def\Cc{{\cal C}}
\def\Dc{{\cal D}}
\def\Ec{{\cal E}}
\def\Fc{{\cal F}}
\def\Gc{{\cal G}}
\def\Hc{{\cal H}}
\def\Ic{{\cal I}}
\def\Jc{{\cal J}}
\def\Kc{{\cal K}}
\def\Lc{{\cal L}}
\def\Mc{{\cal M}}
\def\Nc{{\cal N}}
\def\Oc{{\cal O}}
\def\Pc{{\cal P}}
\def\Qc{{\cal Q}}
\def\Rc{{\cal R}}
\def\Sc{{\cal S}}
\def\Tc{{\cal T}}
\def\Uc{{\cal U}}
\def\Vc{{\cal V}}
\def\Wc{{\cal W}}
\def\Xc{{\cal X}}
\def\Yc{{\cal Y}}
\def\Zc{{\cal Z}}

\def\cross{\!  \times  \!}
\def\alphat{\alpha^{(t)}}
\def\alphab{\alpha^{(b)}}
\def\betat{\beta^{(t)}}
\def\betab{\beta^{(b)}}
\def\ATbf{{\bf A_\sT}}
\def\AHbf{{\bf A_\sH}}
\def\sspan{\mbox{\it span}}

%

%

\twocolumn[

\aistatstitle{Novelty Detection in Time Series via Weak Innovations Representation: A Deep Learning Approach}

\aistatsauthor{Xinyi Wang \And Mei-jen Lee \And Qing Zhao \And Lang Tong}

\aistatsaddress{Cornell University \And Cornell University \And Cornell University \And Cornell University} ]

\begin{abstract}
We consider novelty detection in time series with unknown and nonparametric probability structures. A deep learning approach is proposed to causally extract an innovations sequence consisting of novelty samples statistically independent of all past samples of the time series. A novelty detection algorithm is developed for the online detection of novel changes in the probability structure in the innovations sequence.   A minimax optimality under a Bayes risk measure is established for the proposed novelty detection method, and its robustness and efficacy are demonstrated in experiments using real and synthetic datasets.
\end{abstract}

\section{Introduction}
\label{sec:intro}
Novelty detection is the task of recognizing an unseen pattern in data that deviates from "the norm" defined by historical data or an underlying physical model. This task has broad applications in medical diagnosis, industrial system monitoring, and fraud detection, and it draws techniques from statistics, signal processing, and computer science \citep{MarkouSingh:2003SP,PimentelEtal:2014SP}.

Novelty, by definition, implies that there is no historical record that can be used to quantify or learn its characteristics. To this end, novelty detection can be considered as the challenging special case of anomaly detection when no prior instances or verifiable models of the anomaly can be used.

Novelty detection also differs from outlier detection, although outlier detection techniques are widely applied. In statistics, outliers are samples from the part of the sample space where the occurrence of such samples under the norm is highly unlikely. For novelty detection, in contrast, samples from a "novel" probability distribution may very well be in the part of the sample space with a high probability of occurrence under the norm. 

One of the most difficult challenges in detecting novelty in time series with unknown structures is to account appropriately for the temporal dependencies of time series samples. A standard approach is to take segments of time series and detect novelty in (fixed) finite-dimensional vectors. See, e.g., \cite{MaPerkins:03IJCNN,DasguptaForrest:1995ICIS,GardnerEtal:2006JMLR,DongEtal:2006MSSP}.
Such arbitrary segmentation of the time series may miss temporal dependencies of the time series crucial in novelty detection.   

At a more fundamental level, not knowing the underlying probability models of the time series under the norm and that of the novelty makes it difficult to leverage classical statistical methods such as outlier detections \citep{Ben-Gal2005,BarnettLewis:1994Book} and goodness of fit (GoF) tests \citep{AgostinoStephens:1986Book}. 
Although, in principle, the distribution of normal state  can be learned from data \citep{RobertsLionel:94NC,BarnettLewis:1994Book,Bishop:1994IEECVISP}, for time series with unknown temporal dependencies, nonparametric high-dimensional density learning has prohibitive sample and computation complexities.  
To our best knowledge, no existing computationally tractable techniques can provide some level of performance guarantee in limiting false positive and false negative probabilities when the probability models of the time series under the norm and that of the novelty are both unknown.

In this paper, we take the perspective that novelty samples are from some unknown novelty distribution different from the probability distributions under the norm---herein referred to as the {\em normalcy distribution}, and the problem of novelty detection is to distinguish the arbitrary novelty distribution from that of the norm.  In particular, we consider novelty detection in a less explored problem domain involving time series for which the normalcy probability model of the time series is itself unknown and nonparametric.  We assume instead that the time series data under the norm are available to develop a data-driven nonparametric solution. 

The main contribution of this work is a machine learning approach as a marriage of modern machine learning ideas with two classic mathematical statistic concepts: a generalized (weak) notion of innovations representation of stationary time series by \cite{Wiener:58Book} and goodness-of-fit (GoF) test of \cite{Neyman:1937SAJ}.    Rather than learning the high dimensional (normalcy) distribution, we propose a weak innovation auto-encoder that directly transforms the time series under the norm to a uniform independent and identically distributed ({\it i.i.d.}) {\em innovations sequence}. Such a causal transformation, coupled with Neyman's GoF test, provides an online novelty detector that achieves asymptotic minimax optimality.  We also demonstrate the empirical performance of the proposed novelty detector with real time series measurements collected in a microgrid. 
\vspace{-1em}
\subsection{Related Works}
Novelty detection for time series has been an underdeveloped topic.  One of the earlier novelty detection techniques specifically for time series is the Kalman-filtering-based detection by \cite{LeeRoberts:2008ICPR}, where the innovations sequence of a Gaussian autoregressive time series is extracted.  This particular feature bears considerable similarity to ours, although the technique of Lee \& Roberts is significantly different from ours.  Lee \& Roberts assumed a Gaussian autoregressive model learned using data, which limits the application of their technique.  Our approach, in contrast, learns a nonparametric generative model that produces innovations.

A significant fraction of existing work are based on segmenting the time series into blocks to apply novelty detection techniques designed for independent samples.
In particular, \cite{MaPerkins:03IJCNN} and \cite{GardnerEtal:2006JMLR} proposed to segment time series into vectors and apply One-Class Support Vector Machine (OCSVM) \citep{Scholkopf:99NIPS} to detect novelty.
\cite{DasguptaForrest:1995ICIS} and \cite{DongEtal:2006MSSP} adopted a negative selection technique based on string matching.
They constructed an antibody set representative of the novelty by generating a set of random vectors and removing those matched with normalcy set, judging by their similarity.
Novelty was then detected by measuring the similarity between testing segment and segments in antibody set.
When applied to time series with unknown temporal dependencies, these segmentation-based techniques tend to treat data segments as if they were statistically independent and miss-classify data with unknown temporal dependencies as being novel. 

Our technique presented in this paper takes the general approach of combining representation learning with novelty detection. There have been recent works along this line, including \citep{SchmidtSimic:2019,SteinmannEtal:21} detection problems. For instance, \cite{SchmidtSimic:2019} used the normalizing flow technique to learn an uncorrelated latent representation of the normalcy time series and constructed a likelihood value as the testing statistics.  
\cite{SteinmannEtal:21} adapted variational auto-encoder for representation learning, and combines reconstruction error and Kullback-Leibler distance between representation and standard normal distribution as novelty score for detection. 
Those methods generally don’t ensure the causality of the estimated representation, thus becoming unsuitable for online applications.

A key component of our approach is a Generative Adversarial Network (GAN) approach to learning a causal weak innovations representation of the time series using the auto-encoder structure. Some of the related representation learning can also be used, potentially. For instance, \cite{Brakel&Bengio:17} proposed an autoencoder approach (Anica) to extracting the independent component of the data rather than innovations of the time series. \cite{DinhKruegerBengio:2015} also proposed an independent component extraction method (NICE) via normalizing flow. In \cite{Schlegl&Seebock:19}, an auto-encoder technique is proposed to extract a latent representation using a GAN discriminator. 

\section{Innovations Representation}
\label{sec:bg}
\vspace{-0.5em}
\subsection{Innovations Representation}
In this section, we discuss the concept and history of innovations representation for stationary time series, which reveal its association with time series novelty detection.
\cite{Wiener:58Book} first proposed the concept of {\it innovations sequence}, which is defined to be an invertible {\it i.i.d} sequence $\{\nu_t\}$ causally calculated from the originally observed time series $\{x_t\}$:
\begin{align}
    \nu_t = G(x_t,x_{t-1},\cdots), \label{eq:inn-G} \\
    x_t = H(\nu_t,\nu_{t-1},\cdots), \label{eq:inn-H}\\
    \{\nu_t\}\stackrel{i.i.d}{\sim}\mathcal{U}[0,1]. \label{eq:inn-iid}
\end{align}
Implied from its name, a innovation random variable $\nu_t$ is statistically independent from the past observations $x_{t-1},x_{t-2},\cdots$, and thus represents the new information contained in $x_t$.
The invertible function pair ensures that detecting deviation of $\{\nu_t\}$ from {\it i.i.d} distribution is equivalent to detecting $\{x_t\}$'s deviation from normalcy distribution.
Since the weak innovations representation is calculated casually, it is suitable to be implemented for any real-time applications.
\cite{Kalman:60TASME} and \cite{Kailath:70Proc} have utilized the innovations representation for state estimation as examples.

Although \cite{Rosenblatt:59,Kailath1968TAC,Kalman:60TASME} have found the transformation pair $(G,H)$ for selective parametric models, including Gaussian process, Auto-Regressive (AR) process, Moving Average (MA) process and Markov Chain (MC), no existing work has been done to extract its innovations sequence for general time series.
\cite{WangTong:21JMLR} proposed the first data-driven technique to extract innovations sequence via auto-encoder and Generative Adversarial Networks (GANs), which is referred to as Innovations Auto-encoder (IAE). 
Substantial improvement has been achieved for time series one-class anomaly detection problem through combining IAE with statistical detection method \citep{Paninski:08TIT} developed for uniform distribution. 

Despite its appealing properties, innovations sequence exists under fairly restrictive conditions. 
\cite{Rosenblatt:59} studied necessary and sufficient conditions to extract innovations from general time series, and provided examples for which the innovations sequences don't exist, which includes the widely applied example of two-state Markov chain.
This restricts the capability of the technique being applied to real-world datasets that do not fulfill the existence condition of innovations sequence.
An alternative representation preserving the independence and identity of weak innovations but exists under milder conditions is highly preferable.

\subsection{Weak Innovations Representation}

\cite{Rosenblatt:59} conjectured an {\it i.i.d} representation that can be extracted in the same way as the innovations sequence, from which only a sequence $\{y_t\}$ with the same distribution as $\{x_t\}$ is reconstructed.
Inspired by \cite{Rosenblatt:59}, we propose the utilization of the weak innovations sequence for novelty detection problem, which is mathematically defined through:
\begin{align}
    \nu_t = G(x_t,x_{t-1},\cdots), \label{eq:G1}\\ 
     {\nu_t}\stackrel{i.i.d}{\sim} \mathcal{U}[-1,1],
    \label{eq:G2}\\
    y_t = H(\nu_t,\nu_{t-1},\cdots), \label{eq:H1}\\ 
    \{y_t\} \stackrel{d}{=} \{x_t\}.
    \label{eq:H2}
\end{align}
The weak innovations sequence is defined by relaxing the invertible function pair to "invertible by distribution" defined by Eq.~(\ref{eq:G1}\&\ref{eq:H1}). 
An innovations sequence can be considered as a weak innovations, but not the other direction doesn't hold in general.
For most novelty detection problems, only information with respect to the distribution of $\{x_t\}$ is of interest.
Therefore, intuitively, the representation $\{\nu_t\}$'s capability of reconstructing a sequence $\{y_t\}$ with the same distribution\footnote{This means that for any finite set of time indices $\mathcal{T}$, the joint distribution of $\{x_t\}_{t\in\mathcal{T}}$ is the same as the joint distribution of $\{y_t\}_{t\in\mathcal{T}}$.} as $\{x_t\}$ should be enough for novelty detection problems.

\cite{Rosenblatt:59} and \cite{Wu:05PNAS,Wu11:SI} have discussed the existence condition of the weak innovations sequence.
We summarize their results and form the following theorem explaining the existence condition of weak innovations sequence.
Let the conditional cumulative distribution function (c.d.f.) of a stationary time series $\{x_t\}$ be denoted by:
\begin{multline}
    F(a_t|a_{t-1},a_{t-2},\cdots) = \\ \mathbb{P}[x_t \leq  a_t|x_{t-1}=a_{t-1},x_{t-2}=a_{t-2},\cdots].
\end{multline}
Let $(\mathcal{X},\rho)$ denotes a complete and separable metric space of $x_t, \forall t$, and $\mu(\cdot)$ its probability measure.

\begin{theorem}[Existence Condition of Weak Innovations]\hfill
\vspace{-1em}
    \begin{itemize}
        \item[(a)] A series of random variable satisfying Eq.~(\ref{eq:G1}-\ref{eq:G2}) exists if the conditional c.d.f.'s (as a function of $a_{t}$)  is continuous for almost all\footnote{The almost all statement is with respect to the probability measure of the process $\{x_t\}$} $a_{t-1}, a_{t-2},\cdots$.
        \item[(b)] For time series $\{x_t\}$ that takes the form 
        \[x_t = H'(x_{t-1},\xi_t),\]
        where $\{\xi_t\}$ is a sequence of {\it i.i.d} random variables, $\{x_t\}$ can be written as a causal function of {\it i.i.d} random variable (i.e. Eq.~\eqref{eq:H1}) if 
        \begin{itemize}
            \item[1)] There exist $y_0\in\mathcal{X}$ and $\alpha>0$ such that 
            \[\int \rho^\alpha[y_0,H'(y_0,\xi)]\mu(d\xi)<\infty.\]
            \item[2)] Let $K_\theta^\alpha(\xi) =\sup_{x\neq x'}\frac{\rho[H'(x',\xi),H'(x,\xi)]}{\rho(x,x')}$. There exists $\alpha>0$ such that
            \[\mathbb{E}_{\xi}[K_\theta^\alpha(\xi)]<1.\]
        \end{itemize}
    \end{itemize}
\end{theorem}
This theorem is the direct result of \cite{Rosenblatt:59} and \cite{Wu:05PNAS}, which serve as a sufficient condition for innovations to exist. 
Part (a) of the theorem explains the sufficient condition for Eq.~(\ref{eq:G1}-\ref{eq:G2}) to hold, and part (b) explains the sufficient condition for a time series $\{x_t\}$ to be represented by a causal transform of {\it i.i.d} sequence as in Eq.~\eqref{eq:H1}.

Currently, there is no known techniques to extract weak innovations sequence from time series.
We believe that we made the first attempt to solve open-ended question of extracting the weak innovations sequence through a deep learning approach.

\section{Learning via Weak Innovations Auto-encoder}
\label{sec:learning}
We introduce the Weak Innovations Auto-encoder (WIAE) proposed to learn the weak innovations sequence in Sec.~\ref{subsec:WIAE}. 
The convergence of learned representation to weak innovations sequence when $G,H$ are infinite-dimensional functions are discussed in Sec.~\ref{subsec:Convergence}.  

\subsection{WIAE and Training Objective}
\label{subsec:WIAE}
\begin{figure}[t]
    \centering
    \fontsize{5pt}{1pt}
    \input{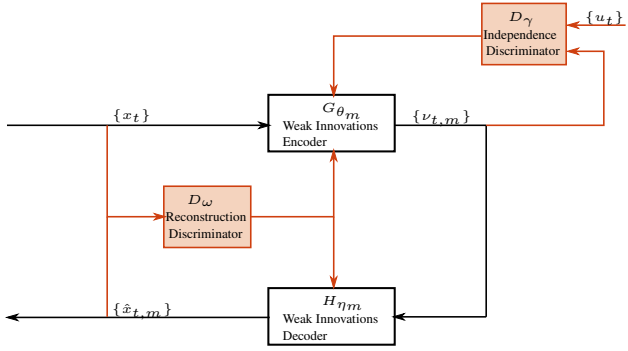}
    \caption{Deep Learning Structure of Weak Innovations Auto-encoder}
    \label{fig:scheme}
\end{figure}

We extract weak innovations sequence through a causal auto-encoder as shown in Fig.~\ref{fig:scheme}. 
The learning framework combines auto-encoder and Wasserstein GAN \citep{Arjovsky17} to fulfill the requirements in Eq.~(\ref{eq:G1}-\ref{eq:H2}). 
The auto-encoder pair, {\it weak innovations Encoder} $G_{\theta_m}$ and {\it weak innovations decoder} $H_{\eta_m}$, are designed to approximate $G$ and $H$.
The auto-encoder pair is parameterized by weights $\theta_m$ and $\eta_m$. 
The subscript $m$ indicates the {\it dimension of the auto-encoders} $G_{\theta_m}$ and $H_{\eta_m}$, which measures the memory size of the auto-encoder pair\footnote{A pair of auto-encoder with dimension $m$ means that its inputs only consist of the past $m$ samples: $(x_t,x_{t-1},\cdots,x_{t-m+1})$.}. 
The causality of the auto-encoder can be guaranteed by adopting the convolutional layers \citep{Waibel&etal:89}.

\def\svgwidth{1\linewidth}

Sequences used for training are reformatted to vectors for compactness.
The vector version is denoted through boldface notations, i.e., $\boldsymbol{G}^{(n)}_{\theta_m}$ and $\boldsymbol{H}^{(n)}_{\eta_m}$. 
The superscripts $n$ represents the {\it dimension of discriminators}, indicating the dimension of output of $\boldsymbol{G}^{(n)}_{\theta_m}$ and $\boldsymbol{H}^{(n)}_{\eta_m}$. 
The output of $\boldsymbol{G}^{(n)}_{\theta_m}$ and $\boldsymbol{H}^{(n)}_{\eta_m}$ were denoted by $\boldsymbol{\nu}_{t,m}^{(n)}$ and $\hat{\boldsymbol{x}}_{t,m}^{(n)}$. 
The two discriminators $\mathbf{D}^{(n)}_{\gamma_m}$ and $\mathbf{D}^{(n)}_{\omega_m}$ measure the Wasserstein distance through Kantorovich-Rubinstein duality \citep{Villani09:Book}, which are then used to train the auto-encoder: 
\vspace{-1em}
\begin{multline}
        \min_{\theta,\eta}\max_{\gamma,\omega}L^{(n)}_m(\theta,\eta,\gamma,\omega) :=\mathbb{E}[\mathbf{D}_\gamma(\boldsymbol{\nu}_{t,m}^{(n)},\boldsymbol{u}_t^{(n)})] \\
        +\lambda\mathbb{E}[\mathbf{D}_\omega(\boldsymbol{\hat{x}}_{t,m}^{(n)},\boldsymbol{x}_t^{(n)})].
    \label{Eq:Criterion}
\end{multline}
Eq.~\eqref{Eq:Criterion} describes the training objective of the weak innovations auto-encoder, where $\boldsymbol{u}_t^{(n)}$ denotes a $n$-dimensional uniform distribution vector where each coordinate is statistically independent. 
The first term in Eq.~\eqref{Eq:Criterion} measures how far $\boldsymbol{\nu}_{t,m}^{(n)}$ is from an {\it i.i.d} uniform sequence, and the second term how well $\hat{\boldsymbol{x}}_t^{(n)}$ reconstructs $\boldsymbol{x}_t^{(n)}$. 
These two discriminators fulfills the requirement of weak innovations representation (Eq.~\eqref{eq:G2} and Eq.~\eqref{eq:H2}) for a $n$-sample block of of learned representation and reconstruction. 
A pseudo-code with detailed training procedure is in the supplementary material.

\subsection{Convergence Analysis}
\label{subsec:Convergence}
It's obvious that the neural network in Fig.~\ref{fig:scheme} is capable of extracting weak innovations by learning $G$ and $H$ when they're finite-dimensional.
For time series in general, the $G,H$ defined in Eq.~(\ref{eq:G1}-\ref{eq:H2}) are infinite-dimensional.
This makes the extraction of weak innovations only possible in an asymptotic sense unless infinite dimensional machine learning framework is practically implementable. 
Thus, in this section we address a structural convergence of WIAE as the dimensionality of neural networks goes to infinite.

Let $\mathcal{A}_m^{(n)}=(G_{\theta^*_m},H_{\eta^*_m})$ denote the finite dimensional auto-encoder pair obtained via optimizing Eq.~\eqref{Eq:Criterion} for $\forall m,n\in\mathbb{Z}^+$. 
Let $\nubf^{(n)*}_{m,t}$ and $\hat{\mathbf{x}}^{(n)*}_{m,t}$ be the weak innovations and reconstruction obtained through $\mathcal{A}_m^{(n)}$, which are reformatted as $n$-dimensional vectors.
Though we would like to have the entire sequence $\{\nu_{t,m}\}\stackrel{d}{\rightarrow}\{u_t\}$ and $\{\hat{x}_{t,m}\}\stackrel{d}{\rightarrow}\{x_t\}$, this cannot be enforced through finite-dimensional training objective.
Therefore, we define the convergence in distribution of finite block, which indicates the convergence in joint distribution of a finite dimension random vector.
\begin{definition}[Convergence in distribution of finite block $n$]
For $n\in\mathbb{Z}^+$ fixed, an $m$-dimensional WIAE $\mathcal{A}^{(n)}_m$ trained with $n$-dimensional discriminators converges in distribution of finite block $n$ to $\mathcal{A} = (G, H)$ if, for all $t$,
\begin{align}
    \nubf^{(n)*}_{t,m} \stackrel{d}{\rightarrow} \nubf^{(n)}_t, \hat{\boldsymbol{x}}^{(n)*}_{t,m} \stackrel{d}{\rightarrow} \boldsymbol{x}^{(n)}_t,
\end{align}
as $m\rightarrow\infty$.
\label{def:convergence}
\end{definition}
Convergence in finite-training block, as defined in Definition.~\ref{def:convergence}, is a compromised version convergence in distribution for infinite series catered to finite dimension implementation of discriminators.
In reality, the training block size $n$ can be chosen with respect to specific requirements and temporal dependency characteristics of certain application.

To achieve convergence of training block $n$ for some $n$ fixed, we make the following assumptions on $\Ac_m^{(n)}$ and $\Ac$:
\ben
\item[A1] {\bf Existence:}  The random process $\{x_t\}$  has a weak innovations representation defined in (\ref{eq:G1} - \ref{eq:H2}), and there exists a causal encoder-decoder pair $(G, H)$ satisfying (\ref{eq:G1}-\ref{eq:H2}) with $H$ being continuous.
\item[A2] {\bf Feasibility:} There exists a sequence of finite-dimensional auto-encoder functions $(G_{\tilde{\theta}_m}, H_{\tilde{\eta}_m})$ that converges uniformly to $(G,H)$ as $m\rightarrow \infty.$
\item[A3] {\bf Training:} The training sample sizes are  infinite. The training algorithm for all finite dimensional WIAE using finite dimensional training samples converges almost surely to the global optimal.
\een

\begin{theorem} 
\label{thm:converge}
Under (A1-A3),  $\Ac_{m}^{(n)}$ converges (of finite block size $n$) to $\Ac$.
\end{theorem}
The proof is in the supplementary material.




\section{Novelty Detection via GoF Tests}
\label{sec:GoF}
\begin{figure}[t]
\center
\includegraphics[scale=0.2]{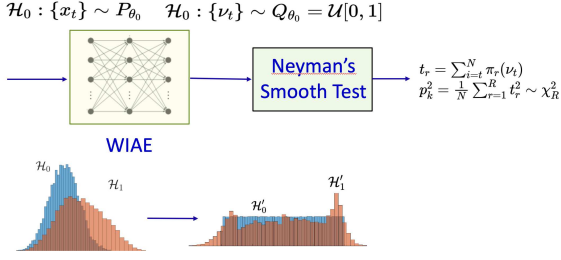}
\caption{\small Novelty detection via WIAE and GoF. Top: Implementation schematic and test statistics under $\Hc_0$.  Bottom left: Histogram of $\{x_t\}$ and $\{\nu_t\}$ under $\Hc_0$ and $\Hc_1$.}
\label{fig:detection}
\end{figure}

In this section, we explained our novelty detection technique based on weak innovations sequence obtained through WIAE.
We then compare the power of novelty detection schemes based on weak innovations $\{\nu_t\}$ and the power of those based on original observation $\{x_t\}$ to show the minimax optimality of our proposed detection method.
We denote the statistical model of $\{x_t\}$ by $\Pc = (X,\mathcal{A},(P_\theta)_{\theta\in\Theta})$\footnote{$\theta$ is the parameter of interest, and it can take any form, i.e. density function, mean, etc.}, and the normalcy distribution by $P_{\theta_0}$.
Given a WIAE trained with data generated from the normalcy distribution $P_{\theta_0}$, we can transform the testing sequences $\{x_t\}$ under $\Pc$ to representations under $\Qc = (Y,\mathcal{B},(Q_\theta)_{\theta\in\Theta})$ through the weak innovations encoder.
The subscript $\theta$ represents the correspondence between the distribution $P_\theta$ of $\{x_t\}$ and the distribution $Q_\theta$ of its learned representation $\{\nu_t\}$. 

We formulate the original time series novelty detection as the following composite hypothesis testing problem:
\begin{multline}
    \mathcal{H}_0: \{x_t\}\sim P_{\theta_0} \text{ v.s. } \mathcal{H}_1: \{x_t\}\sim P_{\theta_1},
    \theta_1\in\Theta_1,
    \label{eq:hypothesis_original}
\end{multline}
where $\Theta_1 = \Theta \setminus \{\theta_0\}$.
Eq.~\eqref{eq:hypothesis_original} defines a challenging but practical formulation of novelty detection, where the inclusion of novelty distribution is maximized.
For general time series whose underlying distribution is not explicitly known, the detection problem is of great difficulty to cope with.

The novelty detection denoted by Eq.~\ref{eq:hypothesis_original} can be transformed to a greatly simplified formulation via WIAE.
For a WIAE trained with normalcy data that follows $P_{\theta_0}$, under ideal conditions specified in Sec.~\ref{sec:learning}, its learned representation should an {\it i.i.d} uniform sequence on $[0,1]$, whose distribution is denoted by $Q_{\theta_0}$.
Hence, the novelty detection base on original $\{x_t\}$ can be transformed to the GoF testing problem of {\it i.i.d} uniform distribution:
\begin{multline}
    \mathcal{H}_0: \{\nu_t\}\sim Q_{\theta_0} \text{ v.s. } \mathcal{H}_1: \{\nu_t\}\sim Q_{\theta_1},
    \theta_1\in\Theta_1.
    \label{eq:hypothesis_innovations}
\end{multline}
Consequently, we can solve the novelty detection problem in Eq.~\eqref{eq:hypothesis_innovations} by leveraging GoF techniques developed for the uniform distributions that measures how well sample data fits $Q_{\theta_0}$. 

Among  various GoF testing techniques for uniform distribution, we use Neyman's smooth test \citep{Neyman:1937SAJ} to conduct novelty detection. 
Neyman designed it to test the null hypothesis of uniform distribution against a class of exponential alternatives. Neyman's smooth test of $R$th order passes a block of testing data through $r$th order Legendre polynomials ($r = 1,2,3,\cdots, R$), and utilizes the summation of square of the output as the testing statistics. 
He also showed that the testing statistics, when scaled properly, has asymptotic $\chi^2$ distribution under $\mathcal{H}_0$.
Further strengthening his results, Neyman proved that the test has the best power among a class of techniques in a neighborhood of null hypothesis, which is also supported through Monte-Carlo analyses \citep{Stephen:1974JASA,QuesenberryMiller:1977JSCS}.
Thus,  we view Neyman's smooth test as an approximation of the locally most powerful test for GoF of uniform distribution, and implemented the test of order $4$ for novelty detection.
The overall procedure of novelty detection is illustrated through Fig.~\ref{fig:detection}.

We evaluate the effectiveness of a novelty detection technique $\phi_\Pc: X\mapsto \{0,1\}$ that takes in samples under statistical model $\Pc$ by its Bayes risk, which is defined as 
\begin{multline}
    R(\Pc,\phi_\Pc,\theta_1) = \mathbb{E}[\phi_\Pc(X)|\theta=\theta_0] \\ +\mathbb{E}[1-\phi_\Pc(X)|\theta = \theta_1],\quad\forall\theta_1\in\Theta_1.
    \label{eq:risk}
\end{multline}
The risk of techniques that takes in samples under statistical model $\Qc$ can be defined similarly, with $\Pc$ changed to $\Qc$. 
A statistical detection procedure is deemed locally optimal in a neighborhood $\Theta'$ around $\theta_0$ if its risk defined by Eq.~\eqref{eq:risk} is the smallest for $\forall\theta_1\in\Theta'$.
We demonstrate the minimax optimality of our novelty detection method by showing the worst-case risk achieved by locally most powerful tests based on $\{x_t\}$ and the locally optimal test based on $\{\nu_t\}$. 
\begin{theorem}
\label{thm:GoF}
\[\sup_{\theta_1\in\Theta_1} \inf_{\phi_\Pc} R(\Pc,\phi_\Pc,\theta_1) = \sup_{\theta_1\in\Theta_1}\inf_{\phi_\Qc} R(\Qc,\phi_\Qc,\theta_1)\]
\end{theorem}
We derive a simple corollary to explain the effect of Theorem.~\ref{thm:GoF}.
\begin{corollary}
\label{coro:GoF}
If for statistical model $\Pc$ and $\Qc$, the locally optimal testing procedure in a neighborhood of $\theta_0$ under Eq.~\eqref{eq:risk} exists (denoted by $\phi_\Pc^*$ and $\phi_\Qc^*$, respectively), then 
\[\sup_{\theta_1\in\Theta'} R(\Pc,\phi_\Pc^*,\theta_1) = \sup_{\theta_1\in\Theta'} R(\Qc,\phi_\Qc^*,\theta_1)\]
\end{corollary}
 Proof for both can be found in the supplementary material.
 The corollary indicates that the locally optimal test conducted based on learned representation $\{\nu_t\}$ achieves the same worst-case (Bayes) risk as the locally optimal test based on original observations $\{x_t\}$.   
Therefore, our novelty detection method of applying locally optimal test (approximated by \cite{Neyman:1937SAJ}'s smooth test) shall also achieve optimal worst-case risk among detection schemes based on $\{x_t\}$.

\section{Numerical Results}
\label{sec:simulation}
\label{subsec:property test}
We evalute both WIAE's capability of extracting weak innovations representation and the power of proposed novelty detection method.
The numerical results of the two parts are presented in Sec.~\ref{subsec:WIAE Performance} and Sec.~\ref{subsec:novelty detection}.

We implemented three other machine learning techniques for comparison. Apart from OCSVM, all techniques aims to extract representation from samples.
\vspace{-1em}
\begin{itemize}
    \item fAnoGAN: fAnoGAN combines GAN and auto-encoder learning to extract the latent representation from original samples, which is proposed by \cite{Schlegl&Seebock:19}.
    Novelty detection was conducted by thresholding on the reconstruction error. 
    
    \item Anica: Anica \citep{Brakel&Bengio:17} is a machine learning technique proposed to learn the independent component in samples.
    The original configuration of Anica doesn't guarantee identity between independent components and the causality of learned representation. 
    To apply Anica to novelty detection, we first converted its learned representation to uniform distribution through empirical c.d.f. and then apply \cite{Neyman:1937SAJ}'s smooth test. 
    
    \item IAE : Proposed by \cite{WangTong:21JMLR}, IAE was designed to extract the innovations sequence $\{\nu_t\}$. The technique has guarantee for extracting innovations sequence when it exists.
    However, when innovations doesn't exist, there is no performance guarantee for IAE.
    Novelty detection based on IAE was conducted by applying \cite{Paninski:08TIT}'s testing method as proposed.
    
    \item OCSVM: OCSVM \citep{Scholkopf:99NIPS} is an unsupervised machine learning technique that learns the decision boundary by controlling the portion of normalcy training sample used as support vectors.
    Since OCSVM doesn't extract representation, it was not included in the evaluation of extracting weak innovations.
\end{itemize}
\vspace{-1em}
We conducted numerical experiments on both synthetic and real datasets.
Three synthetic cases were used for testing as listed in Table.~\ref{tb:Synthetic dataset}. The MA case satisfies the existence condition for weak innovations sequence, while whether it has innovations representation remains agnostic. The Linear Auto Regression (LAR) case on the other hand, has known innovations sequence \citep{WangTong:21JMLR}. Thus the weak innovations sequence must exist for the LAR case. The two-state Markov Chain with transition probability specified in Table.~\ref{tb:Synthetic dataset}, only has weak innovations as shown by \citet{Rosenblatt:59}.  

\begin{table}[t]
\caption{Test Synthetic Datasets. $\nu_t\stackrel{\tiny\rm i.i.d}{\sim}\mathcal{U}[-1,1]$.}
\label{tb:Synthetic dataset}
\begin{center}
\begin{small}
\begin{sc}
\begin{tabular}{ll}
\textbf{Dataset} & \textbf{Model} \\
\hline
Moving Average (MA)    &$x_t=\nu_t+2.5\nu_{t-1}$  \\
Linear Autoregressive (LAR) &$x_t=0.5 x_{t-1}+\nu_t$ \\
Two-State Markov Chain (MC)    &$P=  \begin{bmatrix} 0.6 & 0.4\\0.4 &0.6\end{bmatrix}$\\
\end{tabular}
\end{sc}
\end{small}
\end{center}
\end{table}

Two field data sets were also utilized to test the robustness of the algorithm. These two data sets both contained voltage samples from power systems, and were referred to as UTK and BESS. The BESS data set contains 50 $kHz$ voltage samples collected from a $20kV$ medium voltage micro-grid in EPFL \cite{SossanFabrizio&Namor_2016}. The micro-grid contains a battery that injected a $500kW$ active power into the system, which caused the deviation of voltage signal from the normal state.

The UTK data set was consisted of $6kHz$ voltage data from real-world power system collected by University of Tennessee, Knoxville. There were $180,000$ samples in the data set, with evident high-order harmonics. Only normalcy data were available for the UTK data set, so artificial novelty case was produced by adding Gaussian Mixture noise.

WIAE was implemented with Wasserstein Discriminators as proposed in \citep{Arjovsky17}\footnote{\url{https://keras.io/examples/generative/wgan_gp/}}. All cases were trained with $10,000$ samples. Detailed simulation settings and neural network parameters can be found in the supplementary material.

\begin{table*}[t]
    \centering
    \caption{P-values (p) from Runs Test and Wasserstein Distances (W) $\pm$ Standard Deviation}
    \label{tb:p-value & W-distance}
    \begin{tabular}{ccccccc}
        \textbf{Techniques}  &\textbf{LAR(p)}    &\textbf{MA(p)} &\textbf{MC(p)}   &\textbf{LAR(W)} &\textbf{MA(W)} & \textbf{MC(W)}\\
        \hline
        fAnoGAN     &0.0176 &0.2248 &0.2080 &0.9289 $\pm$ 0.0207 &1.0679 $\pm$ 0.0339 &1.3093 $\pm$ 0.0276\\
        Anica      &$<0.001$   &$<0.001$ &0.0017 &0.8698 $\pm$ 0.0186   &0.8518 $\pm$ 0.074 &0.7112 $\pm$ 0.0231 \\
        IAE    &0.8837 &\textbf{0.9492} &$<0.001$ &0.5555 $\pm$ 0.0551 &0.4839 $\pm$ 0.0202 &0.3984 $\pm$ 0.0687\\
            WIAE   &\textbf{0.9513} &0.8338 &\textbf{0.7651} &\textbf{0.3651 $\pm$ 0.0033} &\textbf{0.3563 $\pm$ 0.0053} &\textbf{0.3814 $\pm$ 0.0053}\\
    \end{tabular}
\end{table*}

\subsection{Performance Evaluation of WIAE}
\label{subsec:WIAE Performance}
We first compare the performance in terms of extracting {\it i.i.d} sequence from original samples. Here we adopted the hypothesis testing formulation know as the Runs Up and Down test \citep{Gibbons:03Book}. Its null hypothesis assumes the sequence being {\it i.i.d}, and the alternative hypothesis assumes the opposite. The test was based on collecting the number of consecutively increasing or decreasing subsequences, and then calculate p-value of the test based on its asymptotic distribution.  According to \citet{Gibbons:03Book}, the runs up and down test had empirically the best performance.

The p-values for runs tests are shown in Table.~\ref{tb:p-value & W-distance}. As seen from the results, only WIAE had the capability of producing seemingly {\it i.i.d} sequences consistently for all cases. Anica, due to its lack of enforcement of identity, didn't perform well under the runs test. fAnoGAN was able to produce the p-value that couldn't allow us to easily reject the null hypothesis for the MA case, but was unable to achieve better p-values for all other cases. IAE performed well for all cases except for the MC case, where the innovations don't exist.
 
The Wasserstein distances between the representation sequence $\{\nu_t\}$ generated by each techniques and an {\it i.i.d} uniform sequence were also calculated to measure the independence of the representation. The Wasserstein distances of reconstruction are presented in Table.~\ref{tb:p-value & W-distance} indicated by $(\mathbf{W})$. The Wasserstein distance were calculated through implementing a Wasserstein discriminator, with the same structure as all other neural networks. WIAE was able to achieve the best Wasserstein distance among all other techniques, with IAE the second best.

\begin{table*}[t]
    \centering
    \caption{ Wasserstein Distances (W) for Reconstruction $\pm$ Standard Deviation}
    \label{tb: Reconstruction}
    \begin{tabular}{cccc}
        \textbf{Techniques}  &\textbf{LAR}   &\textbf{MA} &\textbf{MC}   \\
    \hline
        fAnoGAN      &2.5150$\pm$ 1.0666 &10.9907 $\pm$ 0.2981 &11.5251 $\pm$ 1.7173\\
        Anica       &4.0758 $\pm$ 1.1990   &19.3722 $\pm$ 0.8062 &12.4828 $\pm$ 1.5741 \\
        IAE    &\textbf{1.7119 $\pm$ 0.1110} &2.6697 $\pm$ 0.5221 &2.8264 $\pm$ 0.0996\\
        WIAE    &1.7399 $\pm$ 0.1464 &\textbf{2.0214 $\pm$ 0.1063} &\textbf{1.9593 $\pm$ 0.0338}\\
    \end{tabular}
\end{table*}

\begin{table*}[h]
\caption{Data Detection Test Cases. $\nu_t\stackrel{i.i.d}{\sim}\mathcal{N}(0,1)$,~~$\nu'_t\stackrel{i.i.d}{\sim}\mathcal{U}[-1.5,1.5]$}
\label{tb:Synthetic Test}
\begin{center}
\begin{small}
\begin{sc}
\begin{tabular}{lccl}
\textbf{Test Case} & \textbf{Normalcy Distribution} & \textbf{Novelty Distribution} &\textbf{Block Size} ($N$) \\
\hline
MC1   &$P=  \begin{bmatrix} 0.6 & 0.4\\0.4 &0.6\end{bmatrix}$ &$P=  \begin{bmatrix} 0.62 & 0.38\\0.38 &0.62\end{bmatrix}$ &100\\
AR1    &$x_t=0.5 x_{t-1}+\nu_t$   &$x_t=0.3x_{t-1}+0.3x_{t-2}+\nu_t$  &1000 \\
AR2    &$x_t=0.5 x_{t-1}+\nu_t$    &$x_t=0.5 x_{t-1}+\nu'_t$ &1000\\
MA     &$x_t = \nu'_t + 2.5\nu'_{t-1}$  &$x_t = \nu_t + 2.5\nu_{t-1}$   &500\\
UTK     &Real Data                  &GMM Noise                  &200\\
BESS    &Real Data                  &Real Data                  &500\\
\end{tabular}
\end{sc}
\end{small}
\end{center}
\end{table*}
 
 The Wasserstein distance is shown in Table.~\ref{tb: Reconstruction}.  For the LAR case, since the innovations sequence exists, IAE had the best performance. fAnoGAN was able to achieve reconstruction error close to WIAE for LAR case, yet its extracted components were easily rejected by runs test. For the MC case where only the relaxed innovations sequence exists, WIAE was the only technique able to produce significantly lower reconstruction loss (in terms of Wasserstein distance). Anica was proved to be less competitive in terms of decoded error.

\subsection{Novelty Detection}
\label{subsec:novelty detection}
We tested our novelty detection algorithms under designed novelty cases listed in Table.~\ref{tb:Synthetic Test}. 
The normalcy distributions of synthetic cases stayed the same as the test cases for representation evaluation. 
The AR1 and MC cases tests had designed novelty with the same marginal distribution as the normalcy distribution, while the AR2 and MA had the same temporal dependency structure for the normalcy and novelty. 

In addition to the machine-learning techniques discussed in Sec.~\ref{subsec:property test}, statistical GoF methods for synthetic normalcy distributions were also implemented for their proved effectiveness on parametric models. 
We implemented the Quenouille test \citep{Priestley:81Book} which is designed to test the goodness of fit of an AR($k$) model. 
Due to the asymptotic equivalence between Quenouille's GoF test and the maximum likelihood test \citep{Whittle51:thesis}, we used Quenouille test as a way to calibrate how well WIAE and other nonparametric tests would perform under AR1, for which the Quenouille test is asymptotically optimal. 
Similarly, we used chi-square goodness of fit test for Markov Chain proposed by \cite{Billingsley:1961AnnalsofStats}, and Wold's GoF \citep{Wold:1949RSS} for MA case. 
The performance of each technique was evaluated by comparing its Recipient Operating Characteristics (ROC) curve, which plotted the false positive rate against the true negative rate. 
The Area Under ROC (AUROC) was also calculated accordingly.

\begin{figure}[h]
    \centering
    \includegraphics[scale = 0.4]{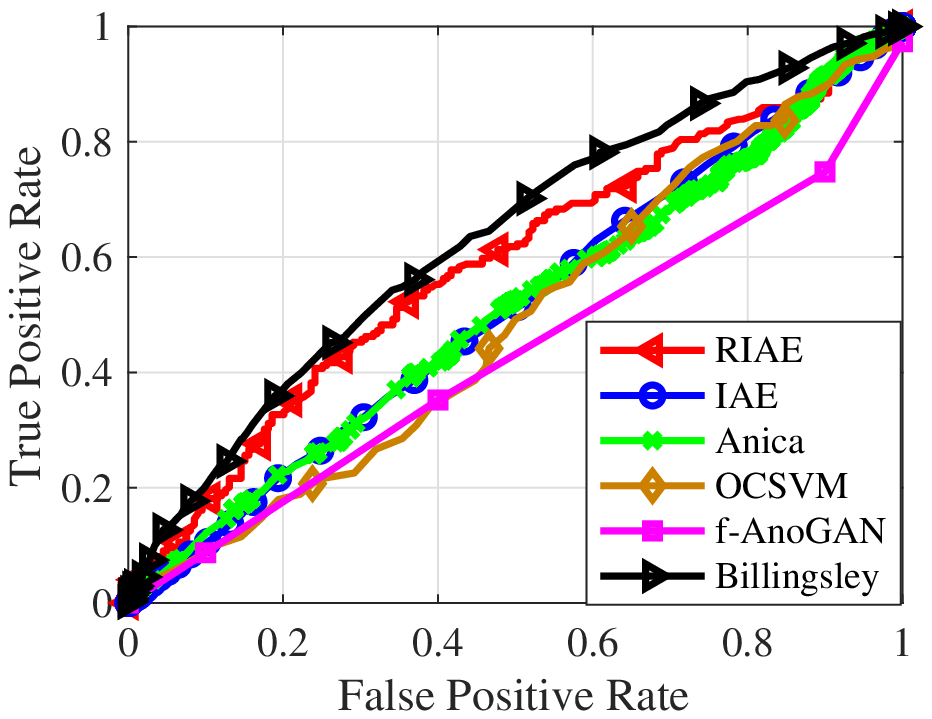}
    \includegraphics[scale = 0.4]{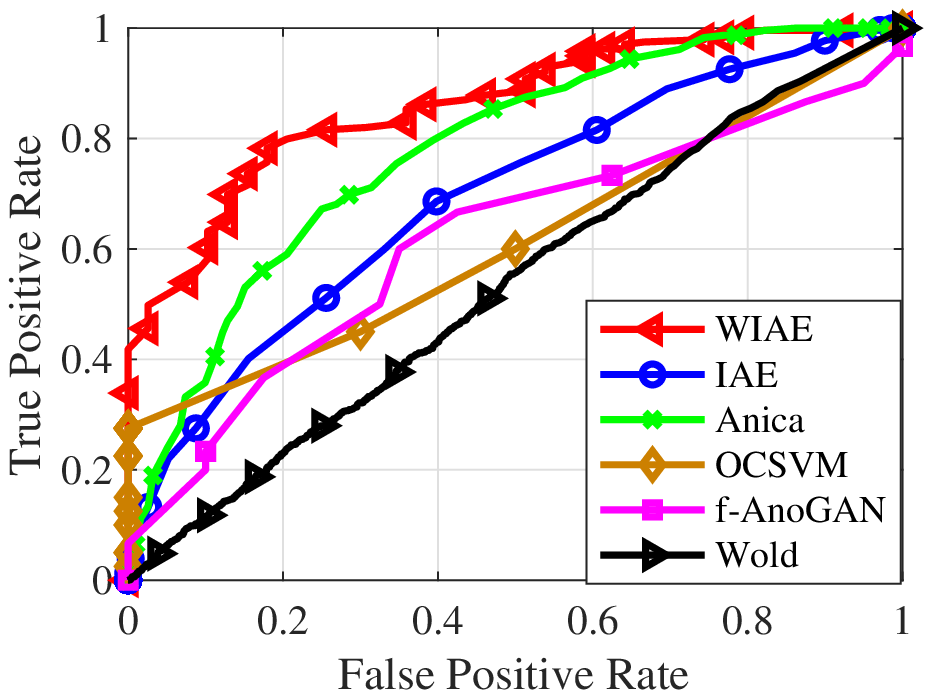}
    \caption{ROC Curves for MC(left) and MA(right). AUROC (left): WIAE: 0.5869 IAE: 0.5118 Anica: 0.2182 fAnoGAN: 0.4541 Billingsley: \textbf{0.6334} AUROC (right): WIAE:\textbf{0.8612} IAE:0.6937 ANICA:0.7786 fAnoGAN:0.6150 Wold:0.5336.}
    \label{fig:ROCs-MA/MC}
\end{figure}

The novelty detection result for MC was shown in Fig.~\ref{fig:ROCs-MA/MC} (left).
To test the robustness of our method, we choose the novelty detection due to its closeness to the normalcy distribution. 
As shown by Fig.~\ref{fig:ROCs-MA/MC}, Billingsley's GoF method developed for MC performed the best.
The result was not surprising since Billingsley's GoF tests is asymptotically equivalent to the generalized likelihood ratio test for Markov chains.
Among all the other schemes, WIAE performed the best, with all the rest performing equally bad.

Fig.~\ref{fig:ROCs-MA/MC} (right) showed the ROC curve for MA case. 
Among all methods, WIAE performed the best. 
Wold's GoF method for MA depends on the auto-correlation, and thus didn't work well for this alternative hypothesis case.
IAE requires the existence of innovations sequence, which is not guaranteed in this case, and had less competitive performance.

\begin{figure}[h]
    \centering
    \includegraphics[scale = 0.4]{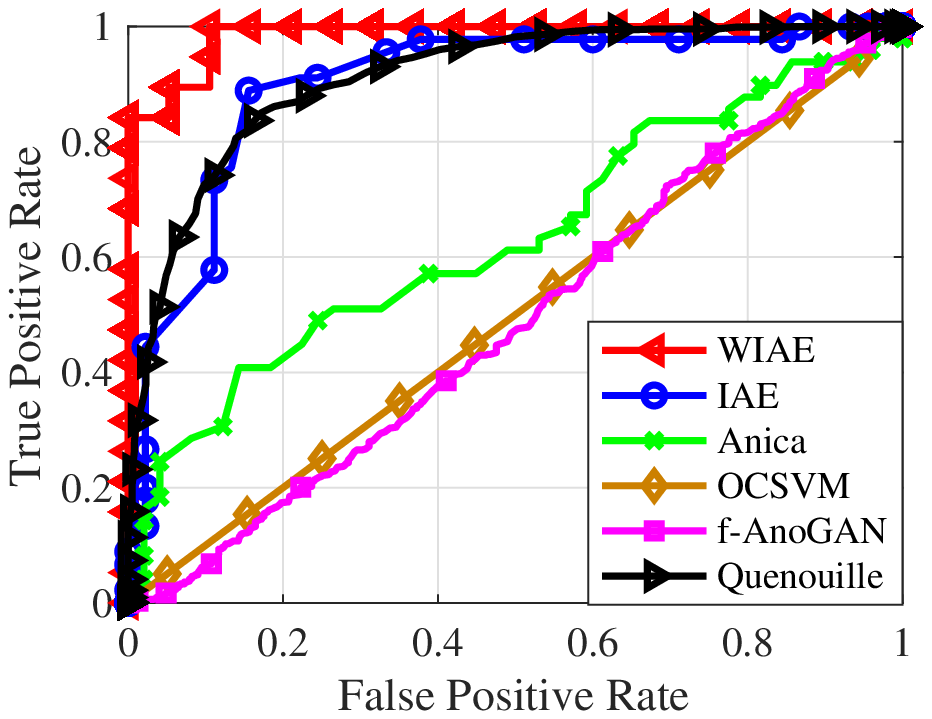}
    \includegraphics[scale = 0.4]{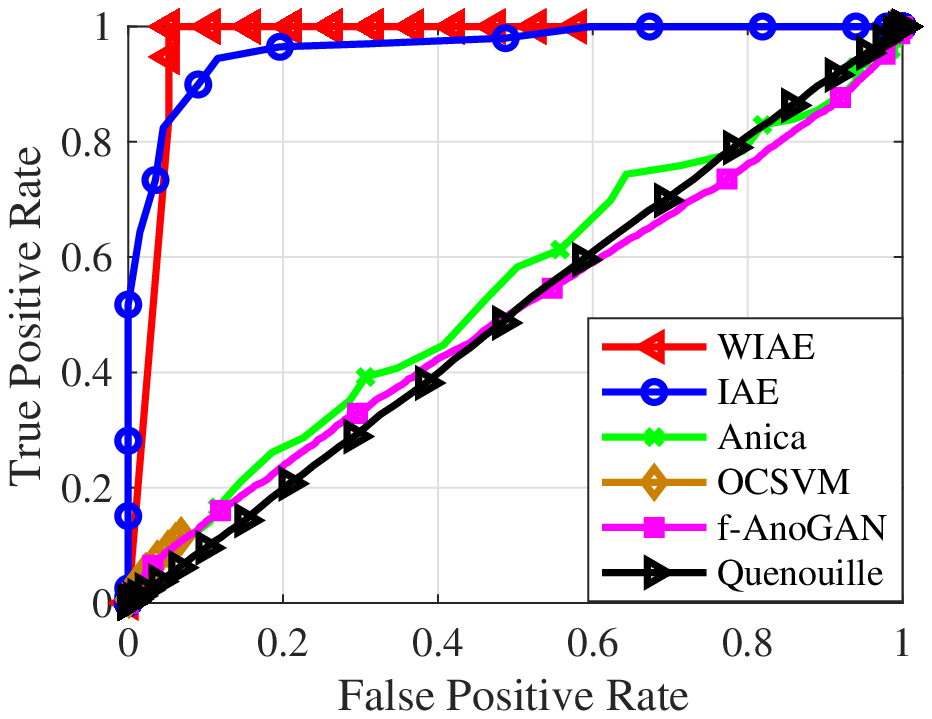}
    \caption{ROC Curves for AR1 (left) and AR2 (right). AUROC(left): WIAE: \textbf{0.9488} IAE: 0.9012 Anica: 0.6337 OCSVM:0.4455 fAnoGAN:0.4881 Quenouille: 0.9112 AUROC (right): WIAE: \textbf{0.9695} IAE: 0.9635 Anica: 0.5407 OCSVM: 0.5407  fAnoGAN: 0.5020 Quenouille: 0.5026}
    \label{fig:ARROCs}
\end{figure}

The ROC curves for the two AR cases are shown in Fig.~\ref{fig:ARROCs}. 
For all AR cases, IAE and WIAE has similar level of performance, since for AR case both innovations and weak innovations exist. 
For AR1 case, IAE achieved similar level of confidence as the asymptotically optimal test, the Quenouille's goodness of fit test, with WIAE performing slightly better. 
We credited the out-performance of WIAE to the easy of its training compared with IAE.
All the other cases fail to perform well under the two synthetic cases, since the support of anomaly and anomaly-free cases largely overlapped.
For AR2 case, since the novelty has the same temporal dependency as the normalcy distribution, Quenouille's GoF tests based on the auto-correlation of normalcy distribution didn't perform well.

\begin{figure}[h]
    \centering
    \includegraphics[scale = 0.4]{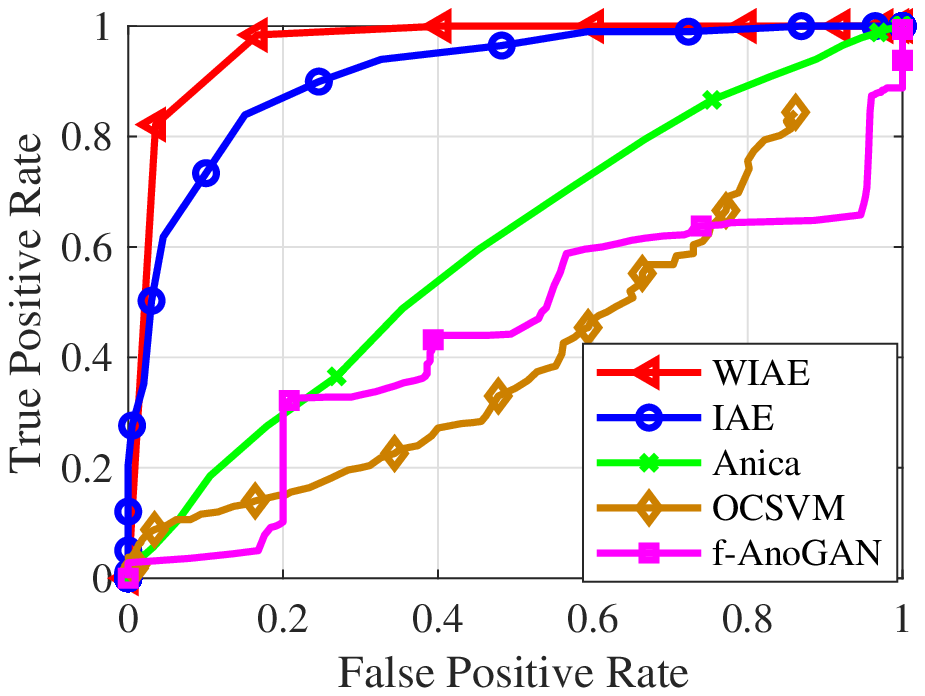}
    \includegraphics[scale = 0.4]{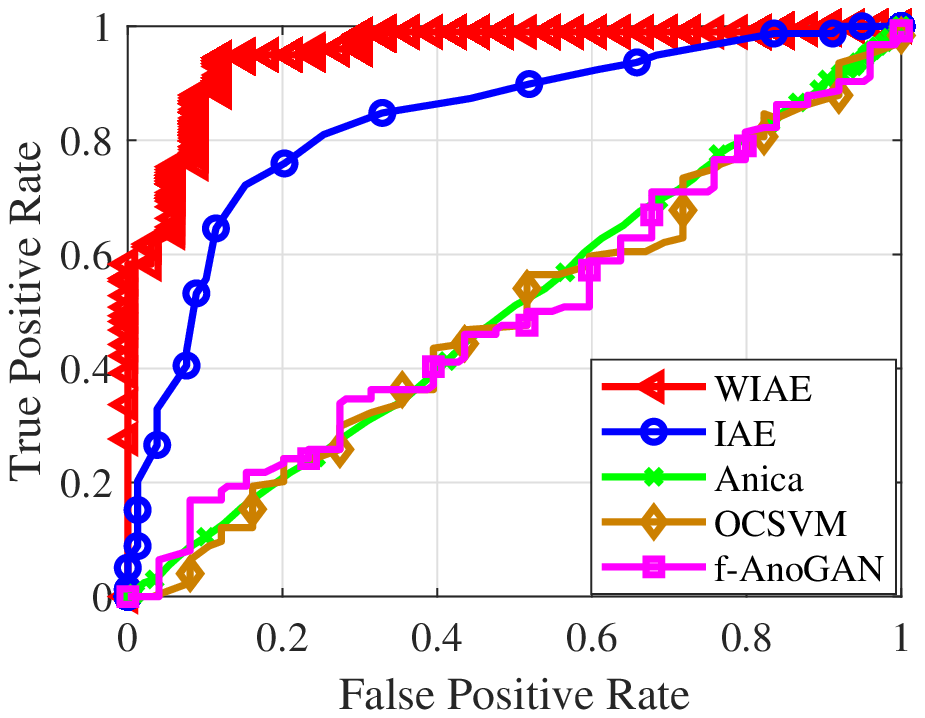}
    \caption{ROC Curves for UTK (left) and BESS (right). AUROC (left): WIAE: \textbf{0.9153} IAE: 0.8642 Anica: 0.5967 fAnoGAN: 0.4394 OCSVM: 0.2987 AUROC (right): WIAE: \textbf{0.9505} IAE: 0.8354 Anica: 0.5027 fAnoGAN: 0.4993 OCSVM: 0.4903}
    \label{fig:RealROCs}
\end{figure}

Fig.~\ref{fig:RealROCs} presents the ROC curves for real datasets. For both cases, IAE and WIAE achieved similar level of performance with WIAE performing slightly better. 
Since the real samples might be generated from a stochastic process which doesn't have innovations representation, WIAE performed better in this case for its superior inclusiveness.

\section{Conclusion}
We provides a deep-learning solution to the challenging non-parametric time series novelty detection problem by combining WIAE and Neyman's smooth test for uniform distributions. 
When trained on normalcy data, WIAE serves as an effective data pre-processor that largely simplifies the novelty detection problem for general time series by converting the novelty detection problem to GoF testing of parametric distribution. 
The proposed novelty detection method achieves minimax optimality under a Bayes risk, and can be readily adapted to real-time applications.
We don't forsee any negative societal impact of the technique so far.
To our best knowledge, we made the first attempt so far to solve the general time series novelty detection problem with optimality guarantee.

\newpage
\begingroup
\raggedright
\bibliography{BIB_RIAE}
\bibliographystyle{icml2022}
\endgroup

\appendix

\onecolumn

\section{Proof of Theorem \ref{thm:converge}}
\label{app:theorem}
For any fixed $n$, we define the following notation:
\begin{align*}
\nu_{t,m}^{*(n)} &= G_{\theta_m^*}(x_t,\cdots,x_{t-m+1}),\\
\hat{x}_{t,m}^{*(n)} &= H_{\eta_m^*}(\nu_{t,m}^{(n)},\cdots,\nu^{(n)}_{t-m+1,m}),\\
\tilde{\nu}_{t,m}^{(n)} &= G_{\tilde{\theta}_m}(x_t,\cdots,x_{t-m+1}).
\end{align*}
We denote the set of weights that obtain optimality under Eq.~\eqref{Eq:Criterion} by $(\theta_m^*,\eta_m^*,\gamma_m^*,\omega_m^*)$. 
We make the clear distinction between the true weak innovations $\{\nu_t\}$ and estimated weak innovations with $m,n$-dimensional Weak Innovations Auto-encoder (WIAE), which is denoted by $\{\nu_{t,m}^{(n)}\}$.

Denote the $n$-dimensional vector $[\nu_{t,m}^{*(n)},\cdots,\nu_{t-n+1,m}^{*(n)}]$ by $\boldsymbol{\nu}_{t,m}^{*(n)}$, with the letter changed to bold face.
Similarly define $\boldsymbol u_{t,m},\hat{\boldsymbol x}_{t,m}^{(n)}$ and $\boldsymbol x^{(n)}_{t,m}$ . $\{\tilde{\nu}_t\}$ is the sequence generated by $G_{\tilde{\theta}_m}$. All the random variables generated by $(\tilde{\theta}_m,\tilde{\eta}_m)$ are defined in the similar pattern.

{\em Proof:} 
\ben
\item Want to show that
$L ^{(n)}(\tilde{\theta}_m,\tilde{\eta}_m,\tilde{\gamma}_m,\tilde{\omega}_m)\rightarrow 0$ as $m\rightarrow \infty$.

By assumption A2, $G_{\tilde{\theta}_m}\rightarrow G$ uniformly, thus $\lVert \tilde{\boldsymbol\nu}_{t,m}^{(n)}-\boldsymbol\nu_t^{(n)}\rVert<n\epsilon$ for $\forall \epsilon>0$. 
Thus $\tilde{\boldsymbol\nu}_{t,m}^{(n)}\stackrel{d}{\Rightarrow}\boldsymbol\nu_t$. 
Similarly, we have $[H_{\tilde{\eta}_m}(\nu_{t},\cdots,\nu_{t-m+1}),\cdots,H_{\tilde{\eta}_m}(\nu_{t-n+1},\cdots,\nu_{t-n-m+2})]\stackrel{d}{\Rightarrow} [x_t,\cdots,x_{t-n+1}]$.

Since $H$ is continuous and $H_{\tilde{\eta}_m}$ converge uniformly to $H$, $H_{\tilde{\eta}_m}$  is continuous. Thus by continuous mapping theorem, $H_{\tilde{\eta}_m}(\tilde{\nu}_{t,m}^{(n)},\cdots,\tilde{\nu}_{t-m+1,m}^{(n)})\stackrel{d}{\Rightarrow}H_{\tilde{\eta}_m}(\nu_{t},\cdots,\nu_{t-m+1})$. Thus, 
\[[H_{\tilde{\eta}_m}(\tilde{\nu}_{t,m}^{(n)},\cdots,\tilde{\nu}_{t-m+1,m}^{(n)}),\cdots,H_{\tilde{\eta}_m}(\tilde{\nu}_{t-n+1,m}^{(n)},\cdots,\tilde{\nu}^{(n)}_{t-n-m+2,m})]\stackrel{d}{\Rightarrow} [x_t,\cdots,x_{t-n+1}]\] 
Hence $L ^{(n)}(\tilde{\theta}_m,\tilde{\eta}_m,\tilde{\gamma}_m,\tilde{\omega}_m)\rightarrow 0$.

Therefore, $L^{(n)}(\theta_m^*,\eta_m^*,\gamma_m^*,\omega_m^*):=\min_{\theta,\eta}\max_{\gamma,\omega}L^{(n)}m(\theta,\eta,\gamma,\omega) \leq L^{(n)}(\tilde{\theta}_m,\tilde{\eta}_m,\tilde{\gamma}_m,\tilde{\omega}_m)\rightarrow 0$.

\item Since $L^{(n)}(\theta_m^*,\eta_m^*,\gamma_m^*,\omega_m^*)\rightarrow 0$ as $m\rightarrow \infty$, by Eq.~\eqref{Eq:Criterion}, $\boldsymbol\nu_{t,m}^{(n)} \stackrel{d}{\Rightarrow} \boldsymbol u_{t,m}^{(n)}$, $\hat{\boldsymbol x}_{t,m}^{(n)}\stackrel{d}{\Rightarrow}\boldsymbol x^{(n)}_{t,m}$ follow directly by the equivalence of convergence in Wasserstein distance and convergence in distribution \citep{Villani09:Book}.
\een
\section{Proof of Theorem \ref{thm:GoF}}
{\em Proof:}
From known relationship between TV distance and testing affinity (i.e., see \citep{pollard05}), we have 
\[\inf_{\phi_\Pc} R(\Pc,\phi_\Pc,\theta_1) = 1-\lVert P_{\theta_0}-P_{\theta_1}\rVert_{TV},\quad \theta_1\in\Theta_1.\] 
By the formulation Eq.~\eqref{eq:hypothesis_original}, we have
\[\sup_{\theta_1\in\Theta_1}\inf_{\phi_\Pc} R(\Pc,\phi_\Pc,\theta_1) = 1-\inf_{\theta_1\in\Theta_1}\lVert P_{\theta_0}-P_{\theta_1}\rVert_{TV}.\]

Since the encoder function $G$ defined by Eq.~\eqref{eq:G1} is measurable, let $G^{-1}(\cdot)$ denote its pre-image, such that $G^{-1}(B):=\{\{x_t\}\in X: G(x_t,x_{t-1},\cdots)\in B\}\in \mathcal{A}, \forall B \in \mathcal{B}$.
Then we have,
\begin{align*}
    \left\lVert Q_{\theta_0}-Q_{\theta_1}\right\rVert_{TV} &= \sup_{B\in\mathcal{B}} \left\lvert Q_{\theta_0}(B)-Q_{\theta_1}(B)\right\rvert = \sup_{B\in\mathcal{B}}\left\lvert P_{\theta_0}(G^{-1}(B))-P_{\theta_1}(G^{-1}(B))\right\rvert\\
    &\leq \lVert P_{\theta_0}-P_{\theta_1}\rVert_{TV}.
\end{align*}

Since $\inf_{\theta_1\in\Theta_1} \lVert P_{\theta_0}-P_{\theta_1}\rVert_{TV}=0$ by the problem formulation,
\begin{align*}
    \left\lvert\sup_{\theta_1\in\Theta_1} \inf_{\phi_\Pc} R(\Pc,\phi_\Pc,\theta_1) - \sup_{\theta_1\in\Theta_1}\inf_{\phi_\Qc} R(\Qc,\phi_\Qc,\theta_1)\right\rvert=
    \left\lvert\inf_{\theta_1\in\Theta_1} \lVert P_{\theta_0}-P_{\theta_1}\rVert_{TV}-\inf_{\theta_1\in\Theta_1}\lVert Q_{\theta_0}-Q_{\theta_1}\rVert_{TV}\right\rvert = 0.
\end{align*}

 \section{Neural Network Parameters}
All the neural networks (encoder, decoder and discriminator) in the paper had three hidden layers, with the 100, 50, 25 neurons respectively. In the paper, $m=20, n=50$ was used for all cases. The encoder and decoder both used hyperbolic tangent activation. The first two layers of the discriminator adopted hyperbolic tangent activation, and the last one linear activation.

In training we use Adam optimizer with $\beta_1=0.9$, $\beta_2=0.999$. Batch size and epoch are set to be $60$ and $100$, respectively. The detailed hyperparameter of choice can be found in Table.~\ref{tb:Hyperparameters}. We note that under different version of packages and different cpus, the result might be different (even if we set random seed). The experimental results are obtained using dependencies specified in Github repository\footnote{\url{https://github.com/Lambelle/WIAE.git}} using a Macbook pro with 2GHz Quad-Core Intel Core i5.

\begin{table}[h]
\caption{Hyper Parameters Setting for Each Case}
\label{tb:Hyperparameters}
\begin{center}
\begin{small}
\begin{sc}
\begin{tabular}{lcccl}
\textbf{Test Case} & \textbf{Learning Rate} $\mathbf{\alpha}$ & \textbf{Gradient Penalty} $\mathbf{\lambda_1,\lambda_2}$ &\textbf{Weight for Reconstruction} ($\mu$) & \textbf{Random Seed} \\
\hline
MC   &$0.0001$ &$1.0,1.0$ &$1.0$ &$140$\\
AR1 \& AR2    &$0.0001$   &$1.0,1.6$ &$1.0$  &$18$ \\
MA     &$0.0001$  &$1.0,1.6$ &$1.0$ &$37$\\
UTK     &$0.0001$  &$1.0,1.2$    &$2.9$   &$80$\\
BESS    &$0.0001$  &$1.0,1.0$    &$1.0$   &$58$\\
\end{tabular}
\end{sc}
\end{small}
\end{center}
\end{table}
 
 \section{Pseudocode}
\begin{algorithm}[h]
  \caption{Training the Equivalent Auto-encoder}
  \label{alg:IGAN}
\begin{algorithmic}
  \STATE {\bfseries Input:} data $(x_t)$, encoder $H_\eta$, generator $G_\theta$, independence discriminator $D_\gamma$, and equivalence discriminator $D_\omega$. $\lambda_1$ and $\lambda_2$ are the gradient penalty coefficients, $\mu$ the weight for auto-encoder, and $\alpha,\beta_1,\beta_2$ hyper-parameters for Adam optimizer.
  \WHILE{Not converged}
  \FOR{$t=1,\cdots,n_{c}$}
        \FOR{$i=1,\cdots,B$}
                \STATE Sample $\xbf_i$ from the input matrix $(x_t)$
                \STATE Sample \textbf{$\mathbf{u}$}$=[u_1,\cdots,u_{n}]^T\stackrel{i.i.d}{\sim}\mathcal{U}[-1,1]$
                \STATE Sample $\epsilon\sim\mathcal{U}[0,1]$
                \STATE $\hat{\nubf}\leftarrow \mathbf{G}_\theta(\mathbf{x}_i)$
                \STATE $\Bar{\nubf}\leftarrow \epsilon \mathbf{u}+(1-\epsilon)\hat{\nubf}$
                \STATE $L^{(i)}_1\leftarrow \mathbf{D}_\gamma(\hat{\nubf})-\mathbf{D}_\gamma(\mathbf{u})+\lambda_1(\lVert \nabla_{\gamma} \mathbf{D}_{\gamma}(\Bar{\nubf})\rVert_2-1)^2$
                \STATE $\hat{\mathbf{x}} \leftarrow \mathbf{H}_\eta(G_\theta(\mathbf{x}_i))$
                \STATE $\bar{\mathbf{x}}\leftarrow\epsilon\mathbf{x}_i + (1-\epsilon)\hat{\mathbf{x}}$
                \STATE $L^{(i)}_2\leftarrow\mathbf{D}_\omega(\hat{\mathbf{x}})-\mathbf{D}_\omega(\mathbf{x}_i)+\lambda_2(\lVert \nabla_{\omega} \mathbf{D}_{\omega}(\Bar{\mathbf{x}})\rVert_2-1)^2$
        \ENDFOR
    \STATE $\gamma\leftarrow Adam(\nabla_\gamma\frac{1}{B}\sum_{i=1}^B L^{(i)}_1,\alpha,\beta_1,\beta_2)$
    \STATE $\omega\leftarrow Adam(\nabla_\omega\frac{1}{B}\sum_{i=1}^B L^{(i)}_2,\alpha,\beta_1,\beta_2)$
    \ENDFOR
    \STATE Sample a batch of $\{\mathbf{x}_i\}_{i=1}^B$ from the input matrix $(x_t)$
    \STATE $\theta\leftarrow Adam\bigg(\nabla_\theta\frac{1}{B}\sum_{i=1}^B\Big[-\mathbf{D}_\gamma(\mathbf{G}_\theta(\mathbf{x}_i))+$
    $\mu\lVert \mathbf{H}_{\eta}(\mathbf{G}_\theta(\mathbf{x}_i))- \mathbf{x}_i\rVert_2\Big],\alpha,\beta_1,\beta_2\bigg)$
    \STATE $\eta\leftarrow Adam\bigg(\nabla_\eta\frac{1}{B}\sum_{i=1}^B\left[\mu\lVert \mathbf{H}_{\eta}(\mathbf{G}_\theta(\mathbf{x}_i))- \mathbf{x}_i\rVert_2\right],$ $\alpha,\beta_1,\beta_2\bigg)$
  \ENDWHILE
\end{algorithmic}
\end{algorithm}


\end{document}